%% file: main.tex
\documentclass[letterpaper, 10 pt, conference]{ieeeconf} 

\IEEEoverridecommandlockouts                              

\usepackage[english]{babel}
\usepackage[utf8]{inputenc}
\usepackage{amsmath,amsfonts,amssymb,bm}
\usepackage{cite}
\usepackage{amsmath,amssymb,amsfonts}
\usepackage{algorithmic}
\usepackage{graphicx}
\usepackage{textcomp}
\usepackage{color}
\usepackage{xcolor}
\usepackage{url}
\usepackage[hidelinks]{hyperref}
\usepackage{booktabs}
\usepackage{multirow}
\usepackage{siunitx}
\usepackage{setspace}
\usepackage{subfig}
\usepackage{bm}
\usepackage{dsfont}
\usepackage{relsize}
\usepackage{soul}
\usepackage[font={footnotesize}]{caption}

\usepackage[ruled,vlined,linesnumbered,boxed]{algorithm2e}

\let\origalg=\algorithm
\def\algorithm{\origalg\setlength{\commentWidth}{7cm}\DontPrintSemicolon\small\SetKwInput{KwParams}{Params}}
\makeatletter
\newcommand{\algrule}[1][.5pt]{\par\vskip.3\baselineskip\hrule height #1\par\vskip.3\baselineskip}
\makeatother
\newlength{\commentWidth}
\newcommand{\subfigureautorefname}[1]{}
\addto\extrasenglish{%
  \renewcommand{\subfigureautorefname}{Fig.}
}

\usepackage{tikz}

\newcommand{\Tmppi}{\ensuremath{N}}
\newcommand{\Kmppi}{\ensuremath{K}}
\newcommand{\lambdamppi}{\ensuremath{\lambda}}

\newcommand{\SO}{\mathbb{S}\mathbb{O}}
\newcommand{\real}{\mathbb{R}}
\newcommand{\x}{\bm{x}}

\renewcommand{\u}{\bm{u}}
\newcommand{\pos}{\bm{p}}
\newcommand{\rot}{\bm{q}}
\newcommand{\vel}{\bm{v}}
\newcommand{\angvel}{\bm{\omega}}
\newcommand{\dt}{\Delta t}
\renewcommand{\S}{\mathcal{S}}

\newcommand{\PREPRINTYEAR}{2024}
\newcommand{\PUBLISHEDIN}{IEEE/RSJ International Conference on Intelligent Robots and Systems (IROS)}
\newcommand{\DOI}{XX.XXXX/XXX.XXXX.XXXXXX} 

\usepackage[placement=top,vshift=-7]{background}
\SetBgScale{1.0}
\SetBgContents{\parbox{\textwidth}{\PUBLISHEDIN. \\ PREPRINT VERSION - DO NOT DISTRIBUTE. \href{https://doi.org/\DOI}{DOI \DOI}}}
\SetBgColor{black}
\SetBgAngle{0}
\SetBgOpacity{1.0}

\title{\LARGE \bf Model Predictive Path Integral Control for Agile Unmanned Aerial Vehicles\\
}

\author{Michal Mina\v{r}\'{\i}k, Robert P\v{e}ni\v{c}ka, Vojt\v ech Von\' asek, and Martin Saska
\thanks{The authors are with the Multi-robot Systems Group, Faculty of Electrical
Engineering, Czech Technical University in Prague, Czech Republic (\protect\url{http://mrs.felk.cvut.cz/}). 
This work  has been supported by the Czech Science Foundation (GAČR) under research project No. 23-06162M, by the European Union under the project Robotics and Advanced Industrial Production (reg. no. CZ.02.01.01/00/22\_008/0004590) and by CTU grant no SGS23/177/OHK3/3T/13.
Computational resources were provided by the e-INFRA CZ project (ID:90254),
supported by the Ministry of Education, Youth and Sports of the Czech Republic. 
}}

\begin{document}

\thispagestyle{empty}
\onecolumn
{
  \topskip0pt
  \vspace*{\fill}
  \centering
  \LARGE{%
    \copyright{} \PREPRINTYEAR~\PUBLISHEDIN\\\vspace{1cm}
    Personal use of this material is permitted.
    Permission from \PUBLISHEDIN~must be obtained for all other uses, in any current or future media, including reprinting or republishing this material for advertising or promotional purposes, creating new collective works, for resale or redistribution to servers or lists, or reuse of any copyrighted component of this work in other works.}
  \vspace*{\fill}
}
\NoBgThispage
\twocolumn
\BgThispage

\maketitle
\thispagestyle{empty}
\pagestyle{empty}

\begin{abstract}
  This paper introduces a control architecture for real-time and onboard control of Unmanned Aerial Vehicles (UAVs) in environments with obstacles using the Model Predictive Path Integral~(MPPI) methodology.
  MPPI allows the use of the full nonlinear model of UAV dynamics and a more general cost function at the cost of~a~high computational demand.
  To run the controller in real-time, the sampling-based optimization is performed in parallel on a graphics processing unit onboard the UAV.
  We propose an approach to the simulation of the nonlinear system which respects low-level constraints, while also able to dynamically handle obstacle avoidance, and prove that our methods are able to run in real-time without the need for external computers.
  The MPPI controller is compared to MPC and SE(3) controllers on the reference tracking task, showing a comparable performance.
  We demonstrate the viability of the proposed method in multiple simulation and real-world experiments, tracking a reference at up to \mbox{44 km h$\mathbf{^{-1}}$} and acceleration close to \mbox{20 m s$\mathbf{^{-2}}$}, while still being able to avoid obstacles.
  To the best of our knowledge, this is the first method to demonstrate an MPPI-based approach in real flight.
\end{abstract}


\section{Introduction} \label{section:introduction}

Agile control of Unmanned Aerial Vehicles (UAVs), commonly referred to as drones, is a challenging problem, mainly when flying in environments cluttered with obstacles as they pose, in general, non-convex spatial constraints.
A typical objective of the control task is to minimize the error between a reference state and the current UAV state~\cite{Tiago2019surveyControl}.
Yet, if the reference is close to the obstacles, the controller should make sure not to collide with the obstacle.
However, the objective highly depends on the application at hand.
In drone racing~\cite{foehn_time-optimal_2021}, the objective is to minimize time or the progress along a reference path~\cite{Romero22MPCCReplanning}.
Nevertheless, even in minimum-time flight through cluttered environments, one might need to combine the time optimality objective with the safety of avoiding obstacles.

The ability to fly in unknown or partially known environments with obstacles is required in many real-world UAV applications, e.g., in
bridge inspection~\cite{Jeong2020BridgeInspection}, inspection of power lines~\cite{ollero2024aerialcore},  aerial coverage scanning~\cite{datsko2024fastCoverage}, or even for digitalization of interiors of historical monuments~\cite{petracek2023ram}.
Among others, the search-and-rescue scenario is the primary motivation for this work as the fast flight can enable the UAVs to reach possible victims faster, even in hard-to-fly environments such as forests or partially collapsed buildings.

\begin{figure}
  \centering
  \includegraphics[width=\linewidth]{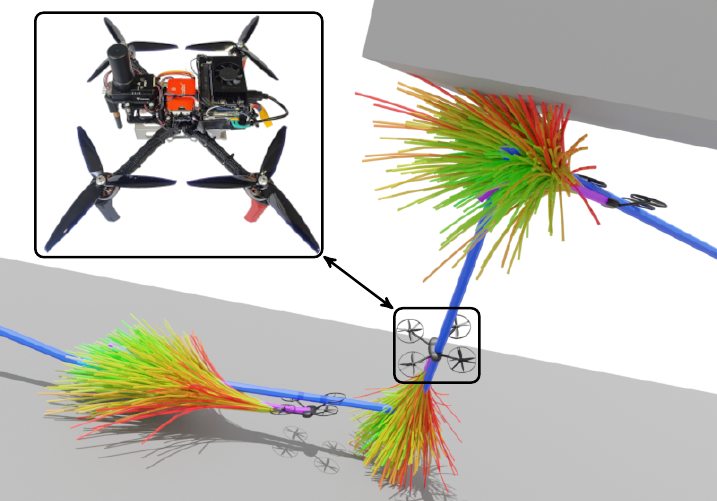}
  \caption{
    The MPPI controller follows a target trajectory (blue) while avoiding collisions with obstacles (gray) by parallel prediction and weighing of hundreds of future trajectories (colored curves).
    Our proposed algorithm is able to run entirely onboard and online.}
  \label{fig:intro}
\end{figure}

The challenges of controlling a UAV in fast and agile flight include modeling obstacles, handling dynamics constraints, and accommodating objectives given by the application while computing the control commands online with high frequency.
This is very demanding for the classical autonomous-flight pipeline, where planning is separated from control.
While trajectory planning accounts for collisions and time allocation of UAV states, it usually does not account for physical constraints such as maximal thrust or angular speeds.
On the other hand, the control usually does not account for obstacle avoidance, which poses a challenge during agile flight, where the deviation from the planned trajectory is typically larger.
Moreover, the controller has to run online with high (e.g., $\SI{100}{\hertz}$) frequency, which is challenging with the increased control problem complexity, such as when considering obstacles or minimum time objectives.

Existing works that can handle various objectives and constraints are primarily based on optimization-based linear~\cite{drones_mpc} or Nonlinear~\cite{recalde_nmpc} Model Predictive Control (NMPC).
The NMPC allows the use of a more precise nonlinear model of the drone dynamics, therefore allowing to constrain the UAV thrust and angular speed limits~\cite{foehn_time-optimal_2021}.
Nevertheless, the collision avoidance constraints are either considered as simplified spherical obstacles~\cite{Lindqvist20NMPCobstacles}, or a limited number of obstacles is modeled~\cite{small2019NPMCobstalces, Garimella17NMPCobstacles} at a control frequency of only 10--20~\SI{}{\hertz}.
However, the other restrictions still hold, such as universal avoidance of non-analytic\footnote{not described by well-known mathematical expressions, such as polynominals and trigonometric functions} and non-convex obstacles, which are almost impossible to introduce into the NMPC framework.
Moreover, the objective of the NMPC also requires to be an analytic function, usually in quadratic form.

In this paper, we use a different predictive control scheme, namely the Model Predictive Path Integral (MPPI)~\cite{williams_aggressive_2016,williams_information-theoretic_2018} originally introduced for car racing.
Using a sampling-based approach that leverages stochastic optimal control, the MPPI removes the restrictions on the cost function, modeled dynamics, and state constraints at the expense of a high computation demand.
This demand is satisfied using the Compute Unified Device Architecture (CUDA) platform, allowing the parallelization of the calculations on a Graphics Processing Unit (GPU).
We propose a method able to track a reference trajectory while using a metric approximation more suitable for UAV rotations.
We show the ability to add an arbitrary collision detection module into the control cost function, thereby including collision avoidance into the control.
We demonstrate all the mentioned abilities on a real-world platform and show that using our approach, it is possible to perform all computations online (at 100~\SI{}{\hertz}) and, more importantly, onboard a real drone while satisfying multiple constraints imposed on the controls and states.
To the best of our knowledge, this is the first MPPI-based approach that can run online on a real drone (compared to existing related works done in simulation) flying fast and agile trajectories.
We show tracking of a reference at speeds of up to $\SI{44}{\kilo\meter\per\hour}$ and acceleration close to $\SI{20}{\meter\per\second\squared}$, even in situations where obstacles are present.

\section{Related Work} \label{section:related_work}

The proposed MPPI controller belongs to model-based control methods that use the UAV dynamics to compute the control commands for a given task.
To this end, we overview the most relevant model-based control approaches~\cite{Tiago2019surveyControl}.

One of the most popular controllers that allow agile flight is the geometric tracking controller~\cite{Lee2011Geo} proposed on the Special Euclidean group SE(3).
The SE(3) controller, also utilized within the MRS UAV Drone system~\cite{Hetrt2023MRS_Drone,baca_mrs_2021}, and later compared to the proposed MPPI approach, is able to avoid singularities commonly associated with Euler angle formulations.
The quadrotor dynamics are shown to be differentially flat~\cite{Mellinger12ijrr}, which enables the computation of both the full state and the control inputs to the system given high-order derivatives of the flat outputs (typically position and heading).
This allows for precise control when the trajectory is given as a high-degree polynomial.
The differential flatness can also be leveraged for control when modeling a multi-rotor UAV with linear rotor drag as shown in~\cite{faessler2017differential}.

One of the most popular and robust control strategies for UAVs is the Model Predictive Control~(MPC)~\cite{drones_mpc}.
Its nonlinear version (NMPC) has even been shown to be able to fly time-optimal agile trajectories designed for drone racing~\cite{foehn_time-optimal_2021}.
A variant of NMPC specially designed for minimum-time flight, called Model Predictive Contouring Control~(MPCC), is capable of computing time allocation of a path during flight~\cite{romero2021model,Romero22MPCCReplanning}.
In~\cite{hanover2021performance}, the authors propose L1-NMPC, which cascades NMPC to an L1 adaptive controller, showing robustness with respect to unmodeled aerodynamic effects, varying payloads, and parameter mismatch.
A comparative study of nonlinear MPC-based and differential-flatness-based control for quadrotor agile flight is shown in~\cite{sun2022comparative}.
While some MPC-based approaches can handle obstacles, they either use only simplified spherical obstacles~\cite{Lindqvist20NMPCobstacles} or a small number of more complex obstacles~ \cite{small2019NPMCobstalces,Garimella17NMPCobstacles}.
At the same time, the control pipeline can run only at a frequency of only 10-20\SI{}{\hertz}.
This prohibits the deployment of the obstacle-aware MPCs to fast flights.

The aforementioned methods are limited in integrating obstacle avoidance.
To cope with obstacles, sampling-based approaches can be utilized.
In~\cite{mbplanner2020}, the authors plan a path for an aerial robot in an environment with obstacles by sampling a set of possible future motion primitives and selecting the best based on task-specific criteria.
Such use of sampling allows general, non-convex and non-differentiable costs, but the result is a high-level path that needs to be followed using additional control algorithms, decoupling the planning and control tasks.
To sample within low-level control sequences, Model Predictive Path Integral can be used, allowing the use of a wide range of cost functions~\cite{williams_aggressive_2016,williams_mppi_2017}, depending on the desired task.

In~\cite{pravitra_mppi_drone_L1}, the authors tackle the task of autonomous drone racing, utilizing a cost function that aggregates 3D centerline following and gate passing, while maintaining the commanded speed and keeping the drone in a given air corridor.
MPPI is used to compute desired collective thrust and body rates, which are then passed to a lower-level controllers.
However, the algorithms are tested in a racing simulation environment with perfect state estimation and the authors do not discuss the hardware requirements needed to sample the numerous future trajectories (7200) at the desired rate (\SI{50}{\hertz}).
Using MPPI to navigate a drone through an environment with obstacles is discussed in~\cite{higgins_mppi_uav_cluttered}.
The authors use MPPI to sample minimum jerk trajectories~\cite{mueller2015minjerk}, whose collision with the environment is then evaluated.
However, the computations are still performed off-board, and the experiments are done in an environment with Vicon.

In this work, we propose an approach allowing real-time and fully onboard agile control of a UAV.
This is achieved by utilizing the MPPI algorithm and taking into account the low-level constraints imposed on the motors directly in the optimization task.
Moreover, we leverage the ability of MPPI to handle more general cost functions, with an important example being universal obstacle avoidance.
Most importantly, we design the rollout prediction and evaluation process with real-world deployment in mind and experimentally prove that the methods are able to run on board a real drone in real-time.
To the best of our knowledge, this is the first MPPI-based used for flying outdoors with a real drone, with all computations running onboard.

\section{Methodology} \label{section:methodology}

\subsection{Model Predictive Path Integral Control} \label{section:mppi}

Model Predictive Path Integral (MPPI) is a predictive control algorithm designed to control nonlinear systems~\cite{williams_aggressive_2016,williams_information-theoretic_2018}.
The core idea is similar to Model Predictive Control (MPC)~\cite{drones_mpc}, however, instead of employing an optimization algorithm, a Monte Carlo sampling approach is used.
This shift to sampling-based optimization allows for general, non-convex cost criteria~\cite{williams_aggressive_2016,williams_mppi_2017}.
Moreover, since no gradient-based optimization is used to find and improve the solution, simple encodings of task descriptions with sparse (or nonexistent) gradients can be utilized.
An example would be testing for a set membership --- yielding a constant in case the robot collides with an obstacle, zero otherwise.
\par
The MPPI optimization algorithm relies on predicting $\Kmppi$ possible future trajectories (\textit{rollouts}) over $\Tmppi$ discrete time samples (\textit{prediction horizon}) in every iteration.
The current state estimate is denoted $\hat{x}$ and the nominal control sequence (sequence of $\Tmppi$ control actions obtained by initialization or previous optimization iterations) as $\bm{u}^{\text{nom}}$.
$\Kmppi$ disturbance sequences of length $\Tmppi$ are sampled from a normal distribution with a zero mean and a covariance matrix $\Sigma$.
We use lower index $j$ to denote the discrete time index and upper index $k$ to denote the rollout index:
\begin{equation}
  \left.\begin{aligned}
    \x^k          & = (\x^k_0, \dots, \x^k_j, \dots,  \x^k_{\Tmppi-1}, \x^k_{\Tmppi})                  \\
    \u^k          & = (\u^k_0, \dots, \u^k_j, \dots,  \u^k_{\Tmppi-1})                                 \\
    \bm{\delta}^k & = (\delta \bm{u}_0^k, \dots, \delta \bm{u}_j^k, \dots, \delta \bm{u}_{\Tmppi-1}^k)
  \end{aligned}\hspace{1em}\right\}\hspace{1em} k = 1, \dots, \Kmppi .
\end{equation}

From the initial state $\hat{x}$, $K$ rollouts are computed by forward simulation of the system dynamics, applying the disturbed nominal control
\begin{equation}
  \left.\begin{aligned}
    \delta \bm{u}_j^k & \in \mathcal{N}(0, \Sigma)                                             \\
    \bm{u}_j^k        & = \bm{u}^{\text{nom}}_j + \delta \bm{u}_j^k                            \\
    \bm{x}_{j+1}^k    & = \bm{x}_{j}^k + \bm{f}_{\text{RK4}}(\bm{x}_j^k, \bm{u}_j^k, \Delta t)
  \end{aligned} \hspace{1em}\right\}\hspace{1em}
  \begin{aligned}
    k & = 1, \dots, \Kmppi       \\
    j & = 0, \dots, \Tmppi - 1 .
  \end{aligned}\label{eq:rollout_computation}
\end{equation}

\begin{figure}[!ht]
  \centering
  \includegraphics[width=\linewidth]{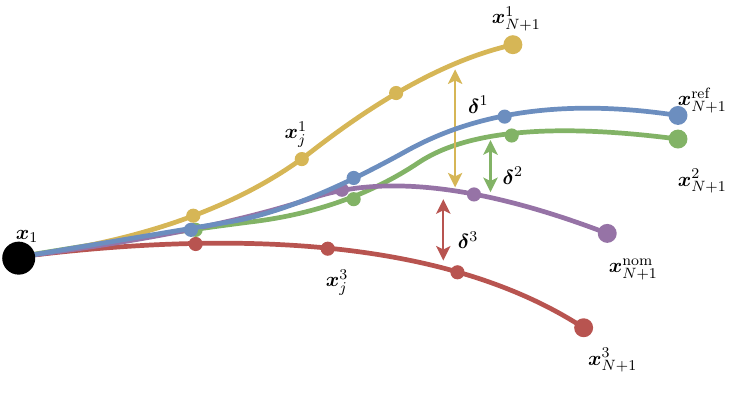}
  \caption{
    To track a given trajectory (blue curve), the MPPI controller keeps nominal control ($\Tmppi$ consecutive control inputs) from previous optimization.
    When applied from the current state $\bm{x}_1$, the nominal rollout is computed (purple).
    The $\Kmppi$ rollouts (other colored curves) are obtained by applying the actions perturbed by $\bm{\delta}^k$ and evaluated with the cost function (green - best, red - worst).}
  \label{fig:evaluation}
\end{figure}

After the rollouts are computed, each rollout is evaluated by a task-specific cost function
\begin{equation}
  \S_k = \text{ComputeCost}(\bm{x}^k, \bm{u}^k).\label{eq:cost_1}
\end{equation}
The lower the cost of the rollout, the better the trajectory is, considering the problem specification (visualized in \autoref{fig:evaluation}).
\par
The costs are then transformed into weights $\left(\omega_1, \omega_2, \dots, \omega_{\Kmppi}\right)$ by
\begin{equation}
  \begin{aligned}
    \omega_k = \frac{1}{\eta}\exp{\left( -\frac{1}{\lambdamppi} \left(\S_k - \rho\right)\right)},   \\
    \eta = \sum_{k=1}^{\Kmppi} \exp{\left( -\frac{1}{\lambdamppi} \left(\S_k - \rho\right)\right)}, \\
    \rho = \min \{\S_1, \dots, \S_{\Kmppi}\}.
  \end{aligned}\label{eq:cost_2}
\end{equation}
This procedure is the softmax transform used often in neural networks to normalize a vector of $K$ values into a probability distribution, where the minimum $\rho$ is subtracted for numerical stability (without changing the result).
\par
After computing the weights, the nominal control actions are updated by a weighted average of the disturbances
\begin{equation}
  \bm{u}^{\text{nom}}_j := \bm{u}^{\text{nom}}_j + \sum_{k=1}^{\Kmppi} \omega_k \cdot \delta \bm{u}_j^k.\label{eq:u_update}
\end{equation}

The parameter $\lambdamppi$ scales the contribution of the trajectory rollouts to the result based on their evaluated cost, ranging from taking only the best rollout to averaging all disturbances.
When $\lambdamppi$ is low, the control disturbances resulting in the best rollout have a significant impact on the updated control.
As $\lambdamppi$ approaches zero, the weight vector approaches $(0, \dots, 0, 1, 0, \dots, 0)$ with one at the place of the best rollouts.
Conversely, when $\lambdamppi$ is high, the weight vector approaches $(\frac{1}{K}, \dots, \frac{1}{K})$.
The whole algorithm is presented in \autoref{alg:MPPI}.
To start the algorithm, we need to initialize the control over the control horizon $\Tmppi$.
In our case, we set the input to zero desired body rates and a constant collective thrust, resulting in a hover state.
\begin{algorithm}
  \setlength{\commentWidth}{2cm}
  \SetKwInput{KwParams}{Params}
  \KwIn{Input initialization $\bm{u}_{\text{init}}$, Problem specific cost \textit{ComputeCost}($\bm{x}^k$, $\bm{u}^k$), System dynamics $\bm{f}_{\text{RK4}}$}
  \KwParams{Number of rollouts $\Kmppi$ and time steps $\Tmppi$, noise covariance $\Sigma$, time step $\Delta t$}
  \algrule
  \For(\tcp*[f]{\makebox[4cm]{Initialize the control\hfill}}){$j$ = 0, \dots, \Tmppi-1}{
  $\bm{u}^{\text{nom}}_j = \bm{u}_{\text{init}}$ \label{alg:MPPI:init1}\;
  }\vspace{0.5em}
  \While{task not completed}{ \nllabel{alg1:while1}
  \setlength{\commentWidth}{4cm}
  $\bm{\hat{x}}$ = \textit{CurrentStateEstimate}() \;
  \For(\tcp*[f]{\makebox[4cm]{Simulate $\Kmppi$ rollouts\hfill}}){$k$ = 1, \dots, $\Kmppi$}{
  $\bm{x}_0^k = \bm{\hat{x}}$ \;
  $\bm{\delta}^k = (\delta \bm{u}_0^k, \dots, \delta \bm{u}_{\Tmppi-1}^k)$, $\delta \bm{u}_j^k \in \mathcal{N}(0, \Sigma)$ \;
  \tcp{$\Tmppi$ steps into the future}
  \For{$j$ = 0, \dots, \Tmppi-1}{
  $\bm{u}_j^k = \bm{u}^{\text{nom}}_j + \delta \bm{u}_j^k $ \;
  $\bm{x}_{j+1}^k$ = $\bm{x}_j^k +  \bm{f}_{\text{RK4}}(\bm{x}_j^k, \bm{u}_j^k, \Delta t)$ \; \label{alg:MPPI:dynamics}
  }
  \tcp{and evaluate their cost}
  $\S_k$ = \textit{ComputeCost}($\bm{x}^k$, $\bm{u}^k$) \label{alg:MPPI:cost}
  }
  \tcp{Compute weights using \autoref{eq:cost_2}}
  $\left(\omega_1, \omega_2, \dots, \omega_{\Kmppi}\right)$ = \textit{ComputeWeights}$\left(\S_1, \S_2, \dots, \S_{\Kmppi}\right)$ \;
  \For{$j$ = 0, \dots, $\Tmppi - 1$}{
  $\bm{u}^{\text{nom}}_j = \bm{u}^{\text{nom}}_j + \sum_{k=1}^{\Kmppi} \omega_k \cdot \delta \bm{u}_j^k$ \tcp{\autoref{eq:u_update}}
  }
  \textit{ApplyToSystem}($\bm{u}^{\text{nom}}_0$) \label{alg:MPPI:apply_to_system}\;
  \For{$j$ = 0, \dots, $\Tmppi-2$}{
  $\bm{u}^{\text{nom}}_j = \bm{u}^{\text{nom}}_{j+1}$ \label{alg:MPPI:shift} \tcp{Shift nominal control}
  }
  $\bm{u}^{\text{nom}}_{\Tmppi - 1}$ = \textit{Initialize}($\bm{u}^{\text{nom}}_{\Tmppi - 1}$)\label{alg:MPPI:init2}\;
  } \nllabel{alg1:while2}
  \caption{Model Predictive Path Integral Control}
  \label{alg:MPPI}
\end{algorithm}

\subsection{Mathematical Model} \label{sec:mathematical_model}
To describe the state of the drone, we use position $\pos \in \real^3$, unit quaternion rotation on the rotation group $\rot \in \SO(3)$ with $\|\rot\| = 1$, velocity $\vel \in \real^3$, and body rates $\angvel \in \real^3$
The evolution of the state $\x = \left[\pos, \rot, \vel, \angvel\right]$ is given by dynamics
\begin{equation}
  \label{eq:drone_dynamics}
  \begin{aligned}
    \dot{\pos}    & = \vel,
                  & \hspace{1em}
    \dot{\rot}    & = \frac{1}{2} \rot \odot \begin{bmatrix} 0 \\ \angvel \end{bmatrix},
    \\
    \dot{\vel}    & = \frac{1}{m} \mathbf{R}(\rot) \begin{bmatrix} 0 \\ 0 \\ F_t \end{bmatrix} + \mathbf{g},
                  &
    \dot{\angvel} & = \mathbf{J}^{-1}\left(\bm{\tau} - \angvel\times\mathbf{J}\angvel\right),
  \end{aligned}
\end{equation}
where $\mathbf{R}(\rot)$ is the matrix representation of the quaternion $\rot$, $\odot$ represents quaternion multiplication, $m$ is the drone's mass, $\mathbf{g}$ is the acceleration due to gravity, and $\bm{J}$ is the drone's inertia matrix.
The control inputs are the torques $\bm{\tau} = \left( \tau_x, \tau_y, \tau_z \right)$ and the collective thrust $F_t$.
The proposed MPPI controller will compute the body rates $\angvel = \left(\omega_x, \omega_y, \omega_z \right)$ and the collective thrust $F_t$, and send them to a lower-level flight controller (described in more detail in \autoref{section:results}).
\pagebreak[4]

Therefore, we introduce \textit{desired body rates} and \textit{collective thrust}
\begin{equation}
  \u_d =
  \begin{bmatrix}
    F_t         \\
    \omega_{xd} \\
    \omega_{yd} \\
    \omega_{zd}
  \end{bmatrix}
  =
  \begin{bmatrix}
    F_t \\
    \angvel_d
  \end{bmatrix},
\end{equation}
used by the MPPI controller as the control inputs $\bm{u}_j^k$.
However, we need to ensure that the commanded change in body rates is feasible (due to motor dynamics, we cannot expect the body rates to change precisely and arbitrarily as we command).
We compute the needed change in body rates over the time step $\Delta t$ as
\begin{equation}
  \dot{\angvel}_d = \frac{1}{\dt} \left( \angvel_d - \angvel \right).
\end{equation}
From $\dot{\angvel}_d $ we can compute the desired body torques
\begin{equation}
  \bm{\tau}_d = \mathbf{J}\dot{\angvel}_d + \angvel\times\mathbf{J}\angvel,
\end{equation}
and the desired single rotor thrusts $ \bm{T}_d$ generating the body torques $\bm{\tau}_d$ and collective thrust $F_t$
\begin{equation}
  \bm{T}_d = \bm{\Gamma}^{-1} \begin{bmatrix}
    F_{t} \\
    \bm{\tau}_d
  \end{bmatrix}.
\end{equation}
The matrix $\bm{\Gamma}$ is the allocation matrix
\begin{equation}
  \bm{\Gamma} =
  \begin{bmatrix}
    1           & 1          & 1           & 1           \\
    -l/\sqrt{2} & l/\sqrt{2} & l/\sqrt{2}  & -l/\sqrt{2} \\
    -l/\sqrt{2} & l/\sqrt{2} & -l/\sqrt{2} & l/\sqrt{2}  \\
    -c_{tf}     & -c_{tf}    & c_{tf}      & c_{tf}
  \end{bmatrix}
\end{equation}
with $l$ being the drone's arm length and $c_{tf}$ being the rotor's torque constant.
The single rotor thrusts $\bm{T}_d$ can be clipped based on the motor constraints $T_{\text{min}}$ and $T_{\text{max}}$ of the drone
\begin{equation}
  \label{eq:clipping}
  \bm{T}_{\text{clip}} = \text{clip}(T_{\text{min}}, \bm{T}_d, T_{\text{max}}).
\end{equation}
The clipped rotor thrusts are then used to compute the clipped body torques and collective thrust
\begin{equation}
  \label{eq:clipped_torque_colthrust}
  \begin{bmatrix}
    F_{t, \text{clip}} \\
    \bm{\tau}_{\text{clip}}
  \end{bmatrix} = \bm{\Gamma} \bm{T}_{\text{clip}}.
\end{equation}
Finally, we reconstruct the feasible inputs as
\begin{align}
  \dot{\angvel}_{\text{clip}} & = \mathbf{J}^{-1}\left(\bm{\tau}_{\text{clip}} - \angvel\times\mathbf{J}\angvel\right), \\
  \angvel_{\text{clip}}       & = \angvel + \dot{\angvel}_{\text{clip}} \cdot \dt.\label{eq:clipped_states}
\end{align}
The clipped values $\left[F_{t, \text{clip}}, \bm{\tau}_{\text{clip}}\right]$ from (\ref{eq:clipped_torque_colthrust}) are used to simulate the dynamics (\ref{eq:drone_dynamics}) and the inputs $\left[F_{t, \text{clip}}, \bm{\angvel}_{\text{clip}}\right]$ from (\ref{eq:clipped_torque_colthrust}) and (\ref{eq:clipped_states}) are sent to the lower-level controller (line \ref{alg:MPPI:apply_to_system} in \autoref{alg:MPPI}).
This allows to consider the limits imposed on the motor thrusts by clipping them accordingly in (\ref{eq:clipping}).

\section{Results} \label{section:results}
The proposed MPPI control algorithm was implemented and integrated into the multirotor Unmanned Aerial Vehicle control and estimation system developed by the Multi-robot Systems Group (MRS) at the Czech Technical University (CTU)~\cite{baca_mrs_2021,Hetrt2023MRS_Drone}.
The experiments were conducted using a high-level computing unit Jetson Orin Nano $\SI{8}{\giga\byte}$ with $\SI{1.5}{\giga\hertz}$ 6-core Arm Cortex-A78AE CPU and $\SI{625}{\mega\hertz}$ 1024-core NVIDIA GPU.
Thanks to the high number of GPU cores, we are able to roll out a large number of parallel simulations (as listed in~\autoref{tab:mppi_parameters}).
In the simulated experiments, we used the multirotor dynamics simulation tools already available in the system~\cite{baca_mrs_2021}.
For the real-world experiments, the computing unit was mounted on a quadrotor based on Readytosky Alien 7" frame with Emax Eco II 2807 $\SI{1700}{Kv}$ motors equipped with 7" three-blade propellers.
The UAV is equipped with Pixhawk Cube Orange+ with PX4 low-level flight controller and Holybro H-RTK F9P GNSS Series RTK GPS receiver for localization.
We use the state estimation directly from the PX4 flight controller connected to the RTK GPS.
The body rates and collective thrust control commands produced by the proposed method are sent directly to the PX4 flight controller that uses body-rate PID to track the commanded body rates.
The reference trajectories are generated using an MPC-based linear trajectory tracker~\cite{baca_mrs_2021}, which produces a smooth receding horizon trajectory that the proposed MPPI controller is tracking.
For safety, additional parameters $\omega_{xy, max}$ and $\omega_{z, max}$ were introduced to limit the desired control applied to the system.
The values used are listed in \autoref{tab:mrs_drone_parameters}.
For an image of the quadrotor, we refer to \autoref{fig:intro}.

\begin{table}[htbp]
  \small
  \centering
  \vspace{0.5em}
  \caption{Drone parameters and MPPI control limits.}
  \vspace{-0.3em}
  \addtolength{\tabcolsep}{-0.2em}
  \begin{tabular}{l r | l r r}
    \toprule
    \multicolumn{2}{c|}{UAV model}                          & \multicolumn{3}{c}{Control limits}                                                                                                                    \\
    Parameter                                               & Value                                                                           & Parameter                                             & Sim  & Real \\
    \midrule
    $m$   \hfill     $[\si{\kilo\gram}]$                    & $1.21$                                                                          & $T_{min}$ \hfill $[\si{\newton}]$                     & 0.3  & 0.3  \\
    $l$    \hfill    $[\si{\meter}]$                        & $0.15$                                                                          & $T_{max}$ \hfill $[\si{\newton}]$                     & 19.0 & 8.0  \\
    $c_{tf}$ \hfill  $[\si{\meter}]$                        & $0.012$                                                                         & $\omega_{xy, max}$ \hfill $[\si{\radian\per\second}]$ & 10.0 & 6.0  \\
    $\mathbf{J}$     \hfill   $[\si{\gram \meter\squared}]$ & diag($\left[\begin{smallmatrix} 7.06 \\ 7.06 \\ 13.6 \end{smallmatrix}\right]$) & $\omega_{z, max}$  \hfill $[\si{\radian\per\second}]$ & 2.0  & 0.5  \\
    \bottomrule
  \end{tabular}
  \label{tab:mrs_drone_parameters}
  \vspace{-2em}
\end{table}

\subsection{Speed of computation}
\begin{figure}[!ht]
  \centering
  \includegraphics[width=\linewidth]{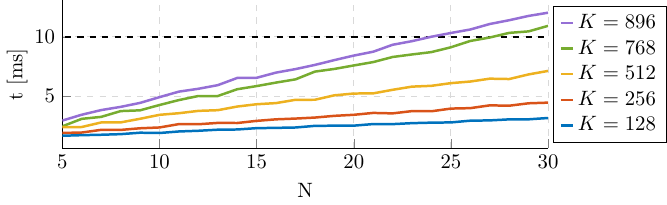}
  \caption{Average time of one MPPI iteration (lines \ref{alg1:while1}--\ref{alg1:while2} in \autoref{alg:MPPI}) depending on the number of rollouts $\Kmppi$ and number of prediction steps $\Tmppi$.
    To run the controller at $\SI{100}{\hertz}$, we need to keep the iteration time below $\SI{10}{\milli\second}$ (black dashed line).
    The computation time may still vary per iteration due to other processes running on the UAV.\vspace{-2em}}
  \label{fig:timing}
\end{figure}

The upper limit for one iteration is $\SI{10}{\milli\second}$ per iteration, which is the update rate of body rate controllers in the~MRS~UAV system ($\SI{100}{\hertz}$).
\autoref{fig:timing} shows how the number of rollouts $\Kmppi$ and the number of prediction steps~$\Tmppi$ affect the iteration time.
In the end, we found the values $\Kmppi = 896$ and $\Tmppi = 15$ to be a good trade-off between the number of trajectories and the prediction steps while making sure the computation finishes in the required time, even when sharing resources with other processes on the UAV.

\subsection{Control input interpolation}
To decouple the time step between two updates of the MPPI controller ($\Delta t = \SI{10}{\milli\second}$ in our case) and the prediction time step, we employ an interpolation scheme.
Decoupling the time steps allows the selection of the prediction horizon independently of the controller frequency.
Otherwise, increasing the prediction horizon would be possible only by increasing $\Tmppi$, however, that would increase the computational complexity, as shown in \autoref{fig:timing}.
We increase the MPPI prediction time step to $n \cdot \Delta t$ and instead of shifting the nominal control by one each time (line \ref{alg:MPPI:shift} in \autoref{alg:MPPI}), we interpolate linearly between them and run the optimization task every iteration.
This allows the use of state feedback at $\SI{100}{\hertz}$, improving the performance of the controller.
In our case, we use $n = 10$, resulting in MPPI having the prediction time step $\SI{0.1}{\second}$.
With $N = 15$ prediction steps, the resulting prediction horizon becomes $\SI{1.5}{\second}$ seconds.

\subsection{Reference tracking} \label{section:reference_tracking}
To design the cost function which assigns a real value to each rollout $\bm{x}^k$ with inputs $\bm{u}^k$, we start with the standard weighing of input and input change
\begin{equation}
  S_k = \sum_{j = 0}^{\Tmppi} \|\u^k_j\|_{R}^2 + \sum_{j = 0}^{\Tmppi - 1} \|\Delta\u^k_j\|_{R_{\Delta}}^2 \label{eq:cost},
\end{equation}
where $R$ and $R_{\Delta}$ are positive semidefinite cost matrices, $\Delta\u^k_j = \u^k_{j+1} - \u^k_j$ denotes the change of input, and \mbox{$\|\cdot\|_{P}^2$} denotes the weighted Euclidean inner product \mbox{$\|\bm{u}\|_{P}^2 = \bm{u}^T P \bm{u}$}.
The states $\x^k_j$ are not penalized, since we only expect to penalize high angular velocities, which can be done by weighing the desired body rates in $\u_j^k$.
\par
\begin{figure}[htbp]
  \centering
  \includegraphics[width=\linewidth]{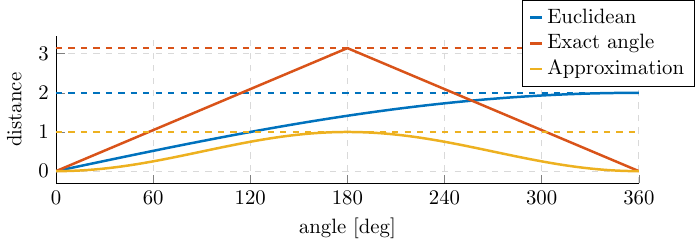}
  \caption{Comparison of the Euclidean distance, exact angle (\autoref{eq:quat_angle}), and approximation thereof (\autoref{eq:quat_approx}).
    The object makes a full 360-degree rotation around a single axis, and the distance to the starting state is computed.
    The Euclidean distance yields a value of $2$ when the object rotates fully, while the angle and its approximation correctly reach maximum at $\SI{180}{\degree}$ and return to zero after the full rotation is completed.}
  \label{fig:quaternion_metrics}
\end{figure}

To enforce tracking the reference, we introduce a term in the form
\begin{equation}
  \sum_{j = 0}^{\Tmppi}\rho_{\text{ref}}(\x^k_j, \x^{\text{ref}}_{j}),
\end{equation}
where $\rho_{\text{ref}}$ is a metric (or at least an approximation thereof) on $\real^3 \times \SO(3) \times \real^3 \times \real^3$
\begin{equation}
  \begin{aligned}
    \rho_{\text{ref}}(\x^k_j, \x^{\text{ref}}_{j}) & = \|\pos^k_j - \pos^{\text{ref}}_{j}\|_{c_{\pos}^{\text{ref}}}^2 + c_{\rot}^{\text{ref}} \cdot d_{\rot}(\rot^k_j, \rot^{\text{ref}}_j)^2                       \\
                                                   & + \|\vel^k_j - \vel^{\text{ref}}_{j}\|_{c_{\vel}^{\text{ref}}}^2 + \|\angvel^k_j - \angvel^{\text{ref}}_{j}\|_{c_{\angvel}^{\text{ref}}}^2, \label{eq:rho_ref}
  \end{aligned}
\end{equation}
with the weighted Euclidean metric used for $\real^3$ and weighing coefficients $c_{\pos}^{\text{ref}}$, $c_{\rot}^{\text{ref}}$, $c_{\vel}^{\text{ref}}$ and $c_{\angvel}^{\text{ref}} \in \real$.
The Euclidean norm is not a proper metric on $\SO(3)$ (quaternion $\rot$ representing the rotation) --- this can be most prominently seen for quaternions $\rot$ and $-\rot$, which represent the same rotation in $\real^3$, but the Euclidean metric will yield a non-zero result.
Therefore, we need to address this inconvenience when adding the reference tracking part to be able to handle distances between quaternions correctly.
There exist two commonly used options for the function \mbox{$d_{\rot} : \SO(3) \times \SO(3) \to \real$}.
The first one is computing the angle of rotation required to get from one orientation to the other, which is given by
\begin{equation}
  \theta = \cos^{-1}(2 \langle \rot_1, \rot_2 \rangle^2 - 1),\label{eq:quat_angle}
\end{equation}
where $\langle \cdot, \cdot \rangle : \SO(3) \times \SO(3) \to \real$ denotes the quaternion inner product
\begin{equation}
  \langle \rot_1, \rot_2 \rangle = w_1w_2+ x_1x_2+ y_1y_2+ z_1z_2.
\end{equation}
However, the evaluation of this function is computationally demanding, and we use instead a common alternative, which roughly corresponds the exact angle (upto a multiplicative constant)
\begin{equation}
  d_{\rot}(\rot_1, \rot_2) = 1 - \langle \rot_1, \rot_2 \rangle^2. \label{eq:quat_approx}
\end{equation}
The advantage of both the exact angle and its approximation over the Euclidean metric is that it better respects the behavior of quaternions, mainly $d_{\rot}(\rot, \rot) = d_{\rot}(\rot, -\rot) = 0$.
The comparison can be seen in \autoref{fig:quaternion_metrics}.

\begin{figure}[htbp]
  \centering
  \subfloat [\textit{line}]{
    \includegraphics[width=0.46\linewidth]{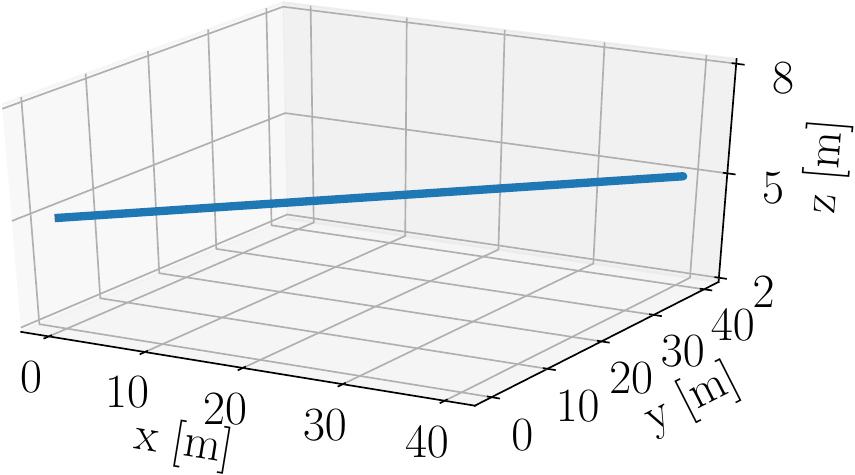}
    \label{fig:traj1}
  }
  \subfloat[\textit{circle}]{
    \includegraphics[width=0.46\linewidth]{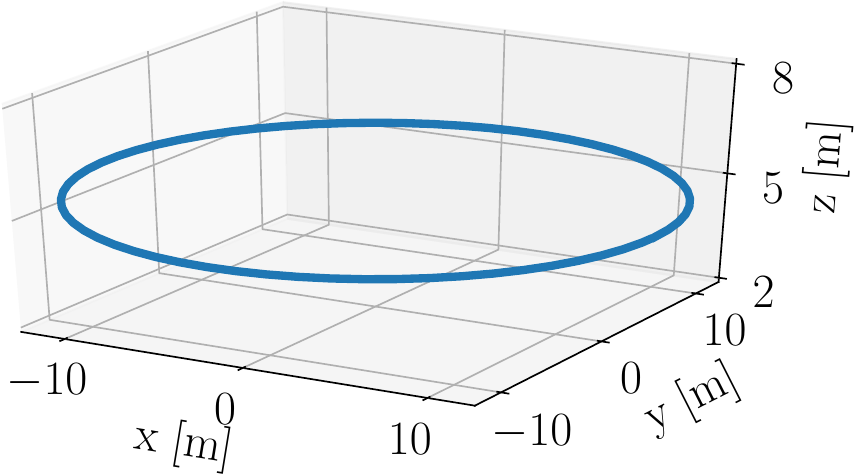}
    \label{fig:traj2}
  }\\\vspace{-0.3em}
  \subfloat[\textit{slanted circle}]{
    \includegraphics[width=0.46\linewidth]{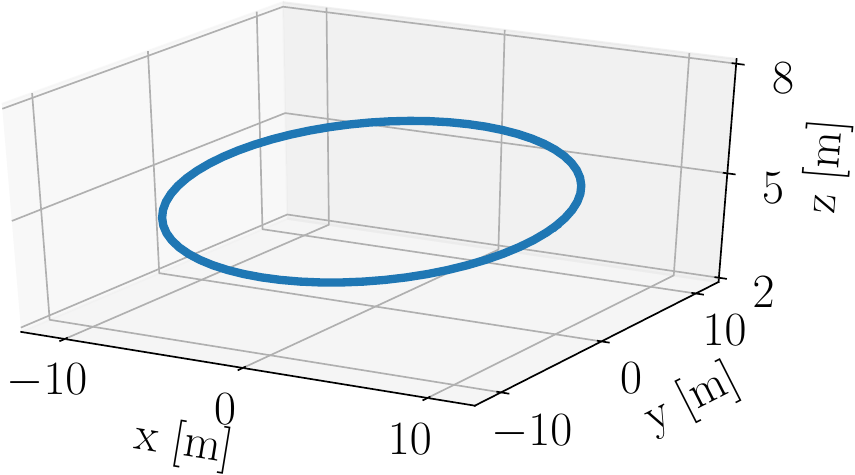}
    \label{fig:traj3}
  }
  \subfloat[\textit{eight}] {
    \includegraphics[width=0.46\linewidth]{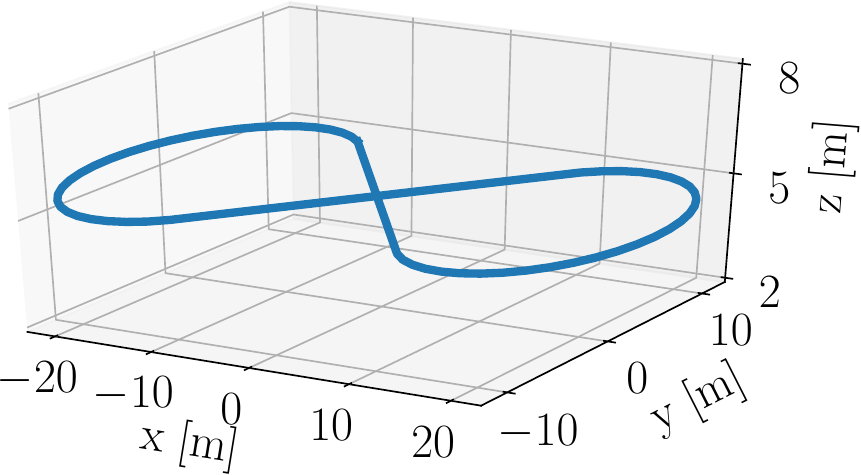}
    \label{fig:traj4}
  }
  \caption{Trajectories used for tracking error comparison.}
  \label{fig:traj}
\end{figure}

\begin{figure}[htbp]
  \centering
  \subfloat [MPPI (\textit{line})] {
    \includegraphics[width=0.47\linewidth]{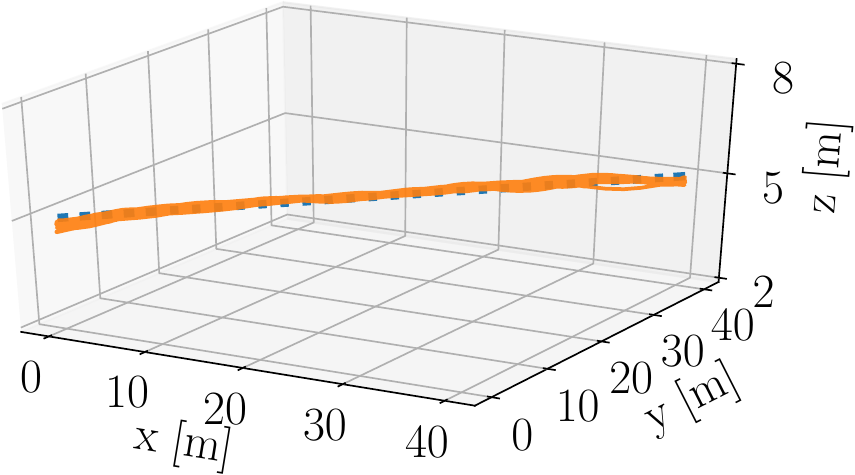}
  }
  \subfloat[MPPI (\textit{eight})]{
    \includegraphics[width=0.47\linewidth]{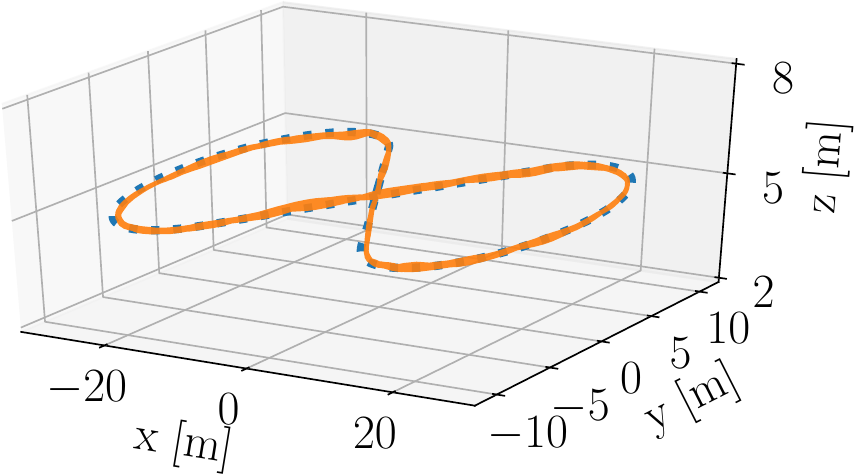}
  }\\\vspace{-0.3em}
  \subfloat[MPC (\textit{line})] {
    \includegraphics[width=0.47\linewidth]{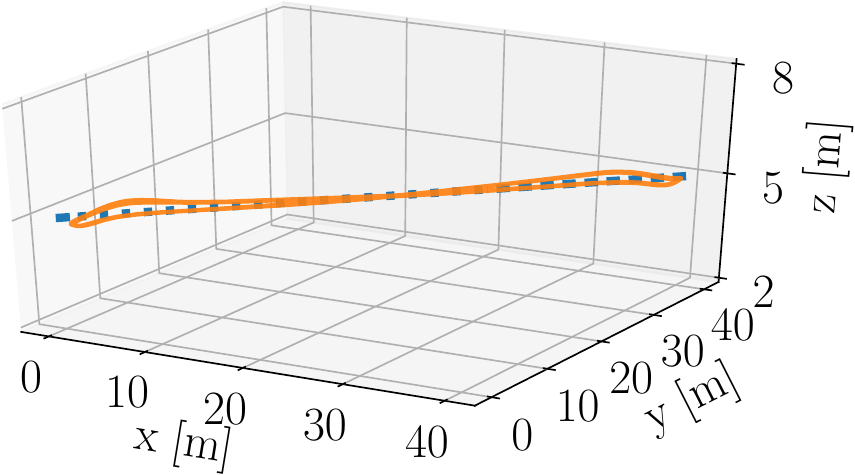}
  }
  \subfloat[MPC (\textit{eight})]{
    \includegraphics[width=0.47\linewidth]{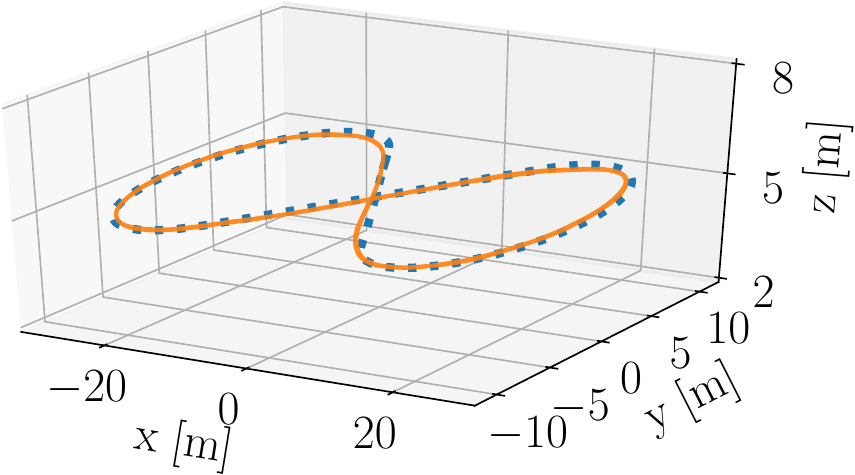}
  }\\\vspace{-0.3em}
  \subfloat[SE(3) (\textit{line})] {
    \includegraphics[width=0.47\linewidth]{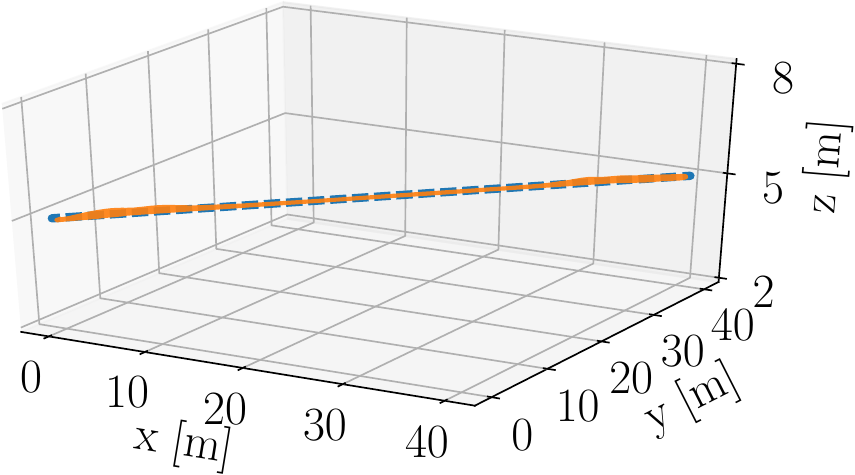}
  }
  \subfloat[SE(3) (\textit{eight})]{
    \includegraphics[width=0.47\linewidth]{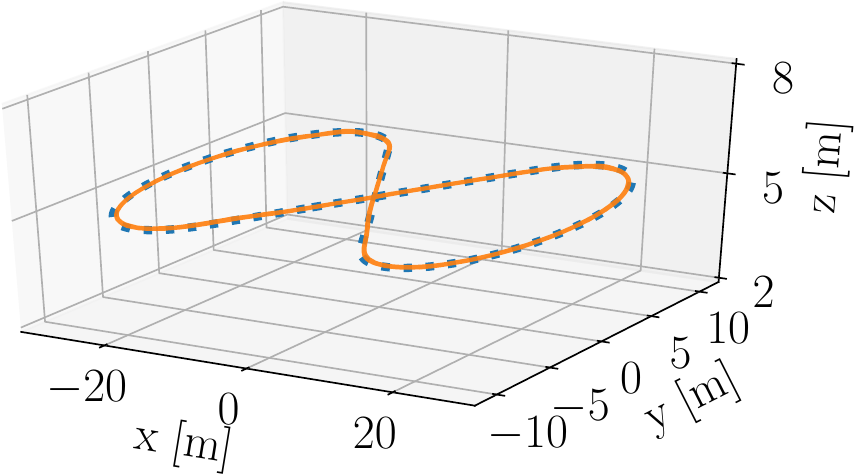}
  }
  \caption{Visualization of the drone trajectories tracked by the MPPI, MPC, and SE(3) controllers in the simulation.}
  \label{fig:tracking_vis_2}
\end{figure}

We generate four reference trajectories (shown in \autoref{fig:traj}) and track them for 20 loops in simulation using the proposed MPPI, MPC~\cite{baca_mrs_2021}, and SE(3)~\cite{Lee2011Geo} controllers.
The MPPI controller parameter values are listed in \autoref{tab:mppi_parameters}, and the resulting tracking errors are presented in \autoref{tab:tracking_error} (with visualizations in \autoref{fig:tracking_vis_2}).
Our controller achieves lower tracking error (around $\SI{60}{\percent}$ improvement) compared to the MPC controller in all cases and reaches results comparable to the SE(3) controller for the \textit{line} and \textit{eight} trajectories.
The SE(3) controller being better than MPPI at the tracking task was expected, since it is specifically made for tracking a trajectory where high-order derivatives of desired states are available.
However, if the derivatives were not available or if the task was more general than simple tracking (e.g., tracking while avoiding obstacles, as presented below), the SE(3) controller cannot be used, whereas MPPI can be used with only minor modifications.
\vspace{-1.5em}
\vspace{-1.2em}
\begin{table}[htbp]
  \small
  \centering
  \caption{Parameter values used in the MPPI controller}
  \vspace{-0.3em}
  \addtolength{\tabcolsep}{-0.1em}
  \begin{tabular}{l r | l r | l r}
    \toprule
    \multicolumn{6}{c}{MPPI parameters}                                                                                              \\
    \midrule
    $\Delta t$    & $\SI{100}{\milli\second}$ & $c_{\pos}^{\text{ref}}$    & $400.0$ & $\Sigma$     & diag($0.60, 0.15, 0.15, 0.05$) \\
    $\Kmppi$      & $896$                     & $c_{\vel}^{\text{ref}}$    & $40.0$  & $R$          & diag($0.01, 0.05, 0.05, 0.10$) \\
    $\Tmppi$      & $15$                      & $c_{\rot}^{\text{ref}}$    & $20.0$  & $R_{\Delta}$ & diag($0.05, 0.10, 0.10, 0.30$) \\
    $\lambdamppi$ & $10^{-4}$                 & $c_{\angvel}^{\text{ref}}$ & $20.0$  &              &                                \\
    \bottomrule
  \end{tabular}
  \label{tab:mppi_parameters}
\end{table}
\begin{table*}
  \vspace{0.5em}
  \caption{Tracking error in simulation (left) and realworld (right) experiments}
  \vspace{-0.3em}
  \centering
  \small
  \begin{tabular}{l c c | c c c | c}
    \toprule
    \multirow{2}{4em}{Trajectory} & $||\bm{v}||_{\text{max}}$  & $||\bm{a}||_{\text{max}}$          & \multicolumn{3}{c|}{Error $[\si{\meter}]$ (\textit{simulation})} & Error $[\si{\meter}]$ (\textit{realworld})                                         \\
                                  & $[\si{\meter\per\second}]$ & $[\si{\meter\per\second\squared}]$ & MPPI                                                             & MPC                                        & SE(3)             & MPPI              \\
    \midrule
    \textit{eight}                & $8.853$                    & $7.571$                            & $0.633 \pm 0.123$                                                & $1.668 \pm 0.396$                          & $0.430 \pm 0.132$ & $0.649 \pm 0.356$ \\
    \textit{slanted circle}       & $5.652$                    & $5.289$                            & $0.412 \pm 0.068$                                                & $0.921 \pm 0.163$                          & $0.061 \pm 0.054$ & $0.518 \pm 0.130$ \\
    \textit{line}                 & $12.271$                   & $19.782$                           & $0.691 \pm 0.310$                                                & $1.704 \pm 0.617$                          & $0.500 \pm 0.198$ & $1.060 \pm 1.148$ \\
    \textit{circle}               & $8.459$                    & $7.034$                            & $0.646 \pm 0.127$                                                & $1.591 \pm 0.202$                          & $0.114 \pm 0.083$ & $0.804 \pm 0.125$ \\
    \bottomrule
  \end{tabular}
  \label{tab:tracking_error}
  \vspace*{-1.5em}
\end{table*}

\begin{figure}[htbp]
  \centering
  \subfloat [\textit{line}]{
    \includegraphics[width=0.47\linewidth]{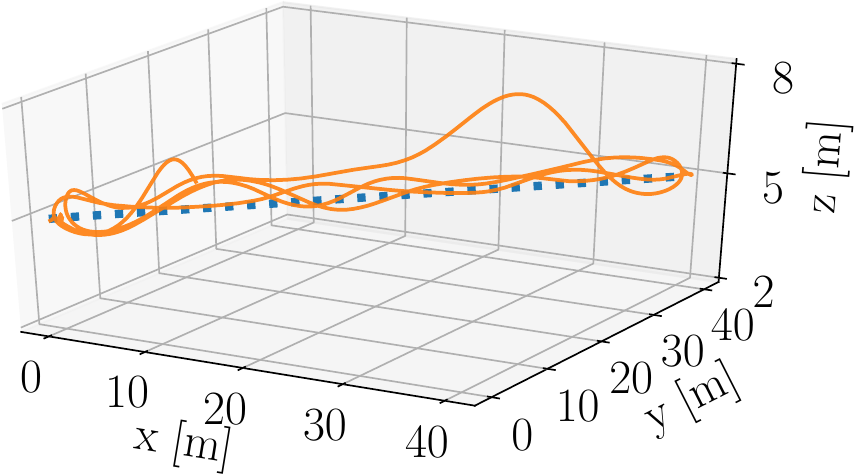}
    \label{fig:tracking_realworld_01}
  }
  \subfloat[\textit{circle}]{
    \includegraphics[width=0.47\linewidth]{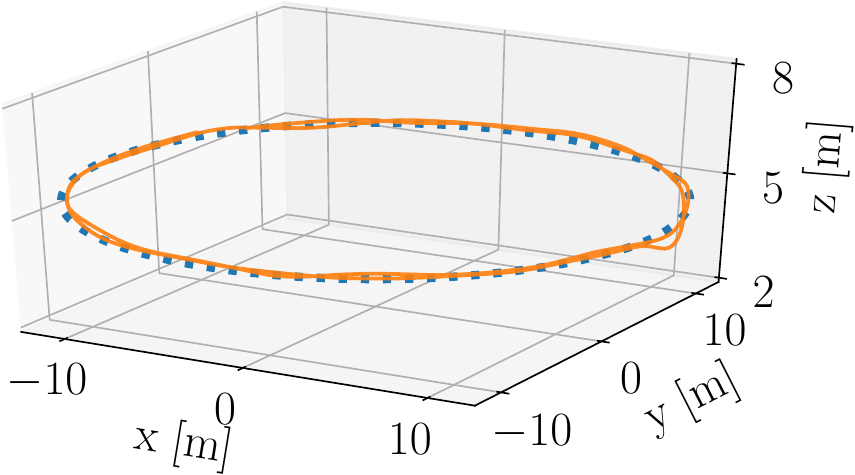}
    \label{fig:tracking_realworld_02}
  }\\\vspace{-0.3em}
  \subfloat[\textit{slanted circle}]{
    \includegraphics[width=0.47\linewidth]{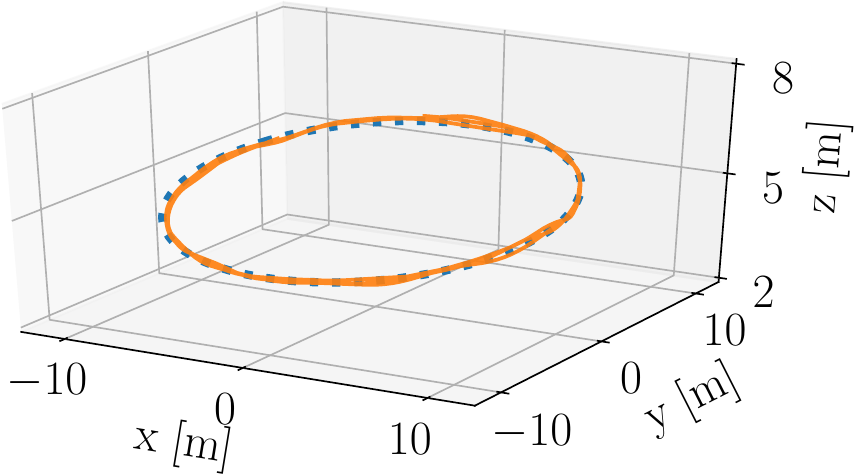}
    \label{fig:tracking_realworld_03}
  }
  \subfloat[\textit{eight}] {
    \includegraphics[width=0.47\linewidth]{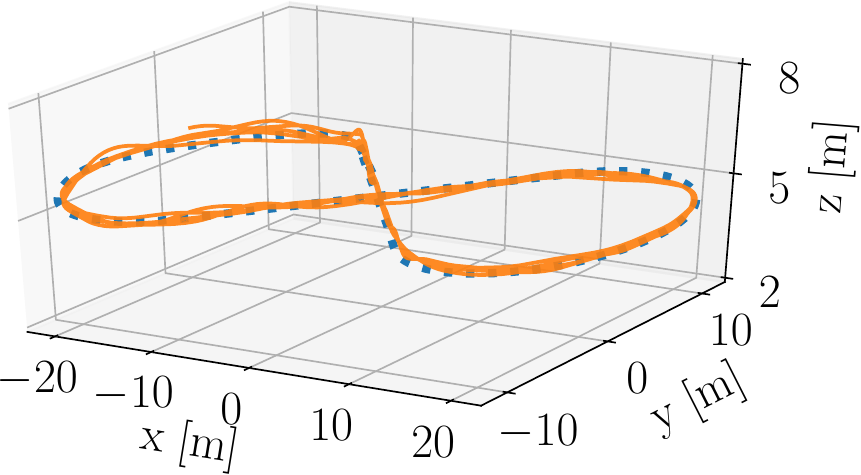}
    \label{fig:tracking_realworld_04}
  }
  \caption{Visualization of the drone trajectories tracked by the MPPI controller in real world.}
  \label{fig:tracking_realworld}
\end{figure}

\begin{figure}
  \centering
  \includegraphics[width=0.95\linewidth, height=0.53\linewidth]{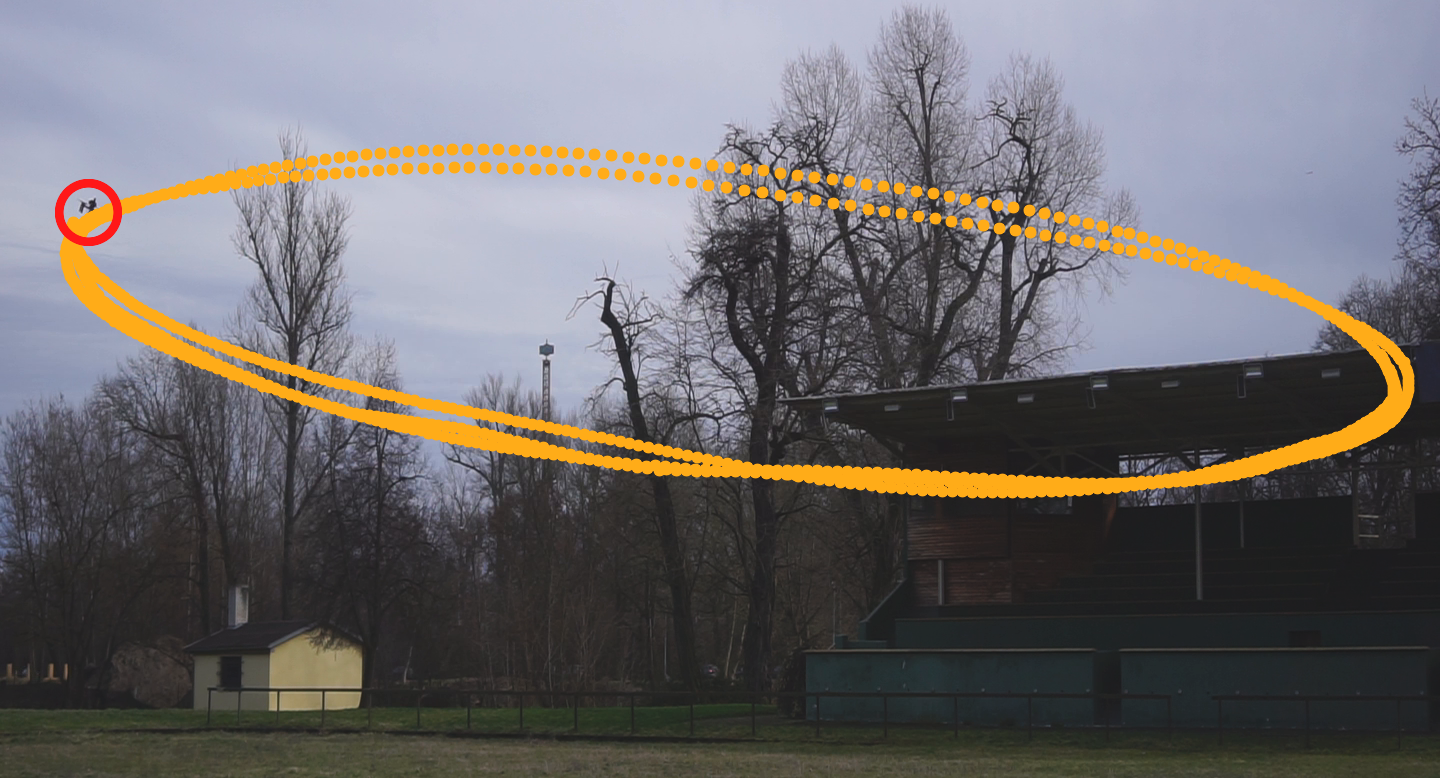}
  \caption{Drone tracked in the video recording of \autoref{fig:tracking_realworld_03}.}
  \label{fig:hw_experiment}
\end{figure}

After verifying our controller in the simulation, we used the same trajectories in real-world validation.
We let our MPPI controller track the four trajectories, each for three laps.
The controller is able to track the trajectories reliably, and the results are comparable to those in simulations, as shown in \autoref{tab:tracking_error} and \autoref{fig:tracking_realworld}.
For the \textit{eight} trajectory, the tracking error is increased by only $\SI{2.52}{\percent}$ when compared to the simulations, which shows a great sim-to-real transfer ability.
On the other hand, the tracking error on \textit{line} is increased by $\SI{53.4}{\percent}$, mainly in the z-axis (\autoref{fig:tracking_realworld_01}).
We believe that this is caused by battery voltage drop at the turn points when under high load, changing the motors' thrust-to-throttle mapping, and could be in the future improved by using a method that accounts for the battery voltage, e.g., by using a voltage-based thrust map.
In \autoref{fig:hw_experiment}, we visualize the drone position history in the video recording of tracking the \textit{slanted circle} trajectory.


\subsection{Obstacles\label{sec:obstacles}}

One of the key advantages of the MPPI method is the ability to include obstacles in the cost function and force the controller to avoid them, which is impossible for the other approaches presented in \autoref{section:reference_tracking}.
This can be achieved by adding a term to the cost function that heavily penalizes states where the drone collides with the environment\vspace{-0.2em}
\begin{equation}
  \S_k^{'} = \S_k + \sum_{j=0}^{\Tmppi} c_{\text{obs}} \cdot \mathlarger{\mathlarger{\mathlarger{\mathds{1}}}}_{\bm{x}_j^k\,\in\,\mathcal{C}_{\text{obs}}},\vspace{-0.2em}
\end{equation}
where $S_k$ is the original cost of the rollout, $c_{\text{obs}}$ is the cost coefficient for the collision term ($c_{\text{obs}} = 10^6$ in our case), and $\mathlarger{\mathlarger{\mathds{1}}}_{\bm{x}_j^k\,\in\,\mathcal{C}_{\text{obs}}}$ is an indicator function, which is $1$ if the drone at state $\bm{x}_j^k$ is in collision with the environment (i.e., $\bm{x}_j^k$ lies in the obstacle region $\mathcal{C}_{\text{obs}}$ determined by a collision detection module).
Collision detection can be implemented in multiple ways (e.g., using mesh collision detection such as \textit{Rapid} \cite{gottschalk_rapid} or discrete grid-based methods, such as \textit{OctoMap} \cite{hornung13auro}), and the exact implementation depends on the platform capabilities (e.g., Light Detection and Ranging (LiDAR)).
\par
\begin{figure}[!htbp]
  \vspace{-1em}
  \centering
  \subfloat {
    \includegraphics[width=0.21\linewidth]{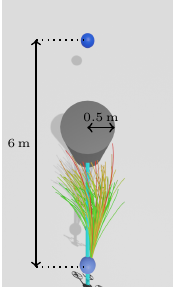}
    \label{fig:mppi_obstacle_01}
  }
  \subfloat {
    \includegraphics[width=0.21\linewidth]{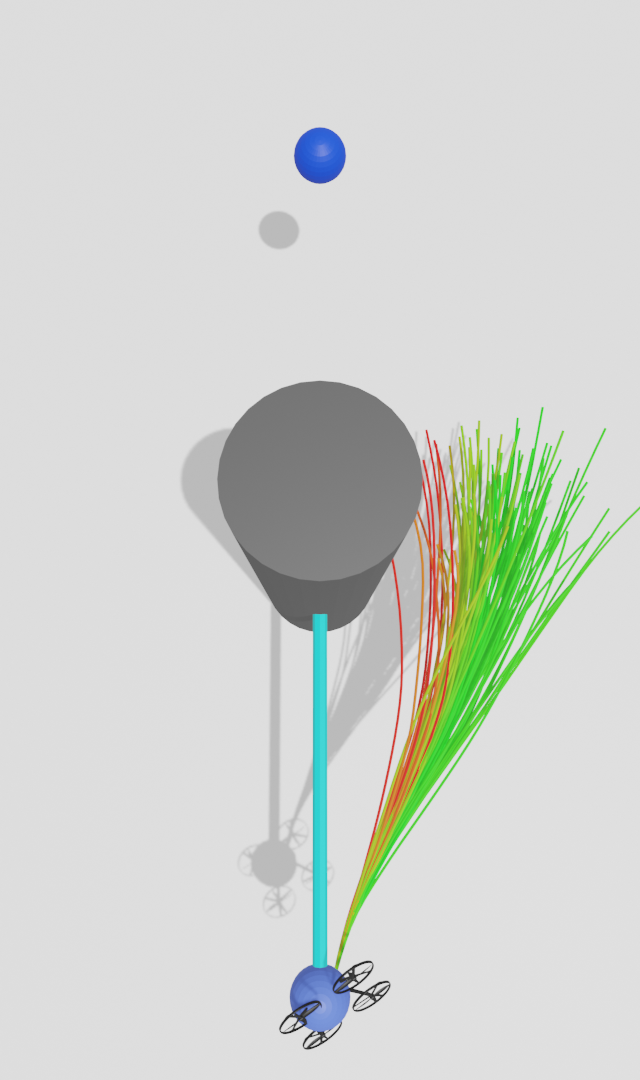}
    \label{fig:mppi_obstacle_02}
  }
  \subfloat {
    \includegraphics[width=0.21\linewidth]{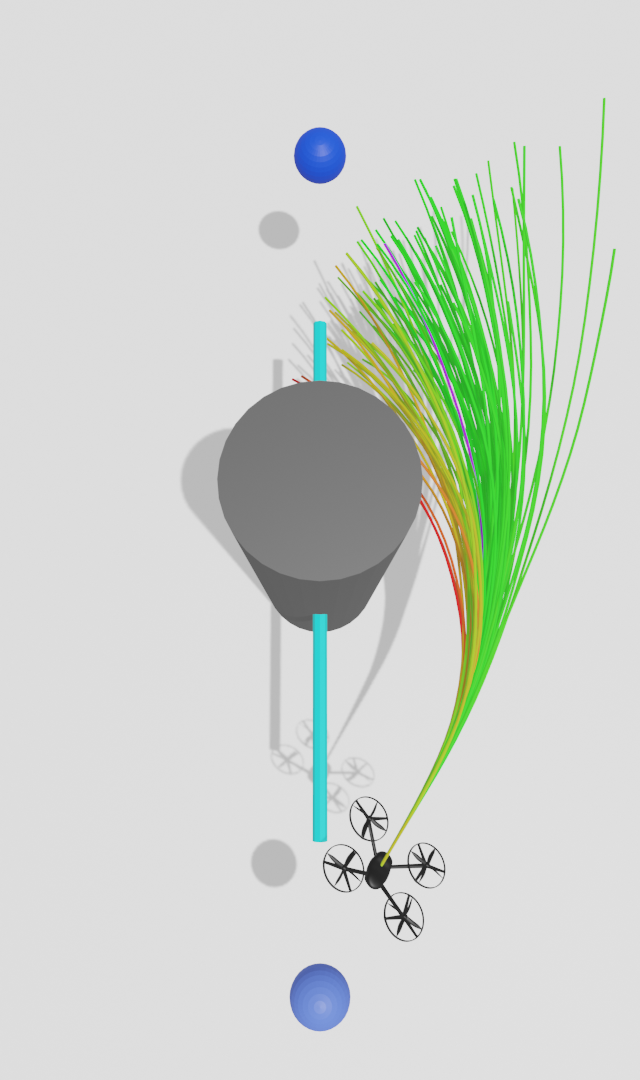}
    \label{fig:mppi_obstacle_03}
  }
  \subfloat {
    \includegraphics[width=0.21\linewidth]{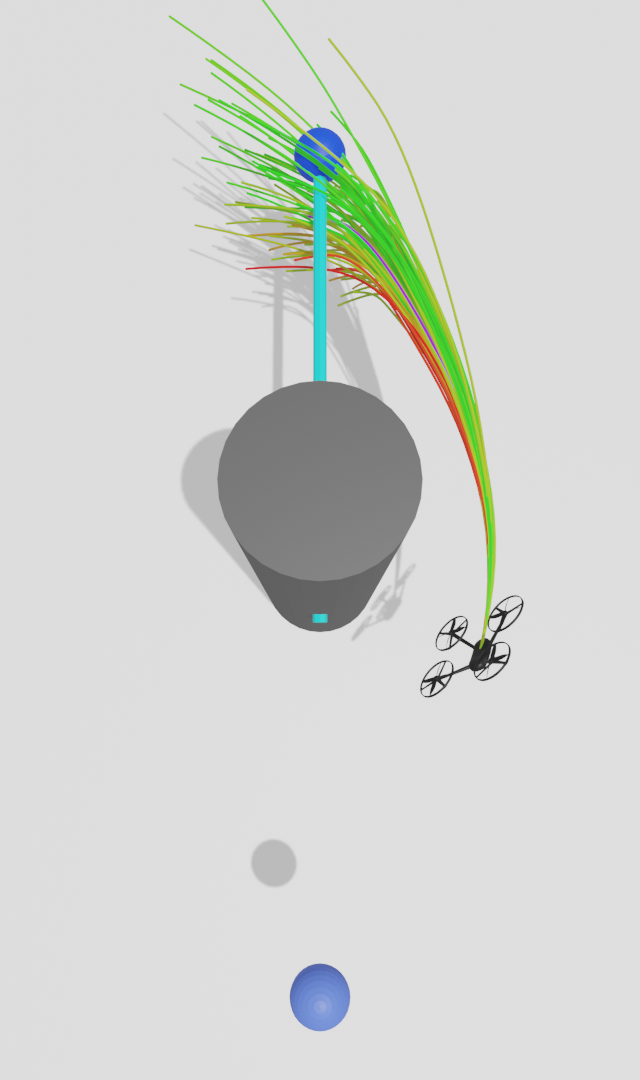}
    \label{fig:mppi_obstacle_04}
  }
  \caption{
    Tracking a straight line moving between the blue spheres at a constant speed of $\SI{6}{\meter\per\second}$ (cyan), while avoiding an obstacle (gray cylinder).
    The colliding rollouts are heavily penalized (red curves) in comparison to the non-colliding ones (green).\vspace{-2em}
  }
  \label{fig:mppi_obstacle}
\end{figure}
\begin{figure}[htbp]
  \subfloat [\textit{slanted circle}]{
    \includegraphics[height=0.42\linewidth, trim=0 0 -0.5cm 0]{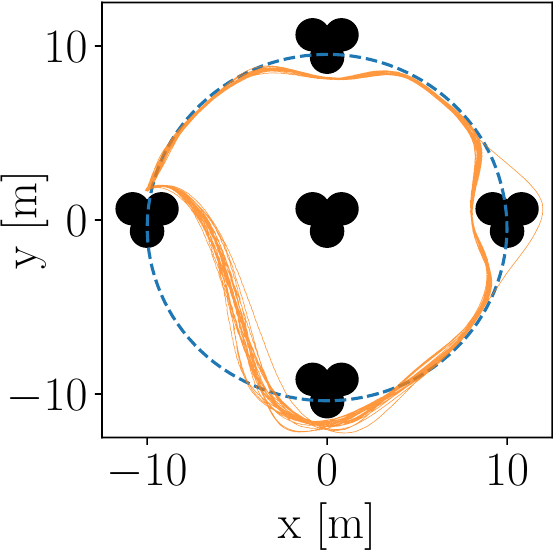}
    \label{fig:sim_obstacle_01}
  }
  \subfloat [\textit{line}]{
    \includegraphics[height=0.42\linewidth, trim=0 0 -0.5cm 0]{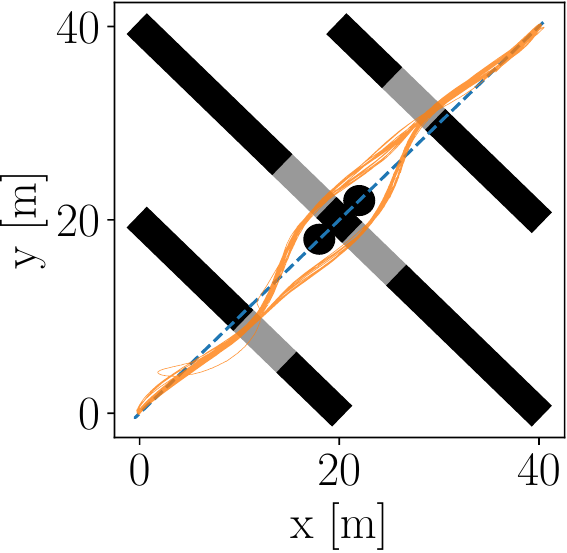}
    \label{fig:sim_obstacle_02}
  }
  \caption{
    We test our MPPI algorithm on \textit{slanted circle} (\autoref{fig:traj3}) with \mbox{pillar-like} obstacles (black) and on \textit{line} (\autoref{fig:traj1}) with walls (black, windows in gray) and let the drone loop the desired trajectory (blue curve) 30 times.
    The recorded UAV positions are shown in orange.\vspace{-2.5em}}
  \label{fig:sim_obstacle}
\end{figure}

First, we validate the ability to avoid obstacles in an environment with a single cylindrical pillar, shown in \autoref{fig:mppi_obstacle}.
Further, we add virtual obstacles to two of the tracked trajectories (\textit{line} and \textit{slanted circle}) and task the controller to track the trajectories 30 times in a row in the simulation, keeping all of the parameters as in \autoref{section:reference_tracking}.
In \autoref{fig:sim_obstacle} we show that the MPPI controller is able to handle and avoid the obstacles, successfully finishing the tracking task without any collisions.
It is even able to consistently cope with non-convex obstacles and situations where the drone needs to retreat after being led to a dead-end, as is the case near the leftmost obstacle in \autoref{fig:sim_obstacle_01}.
Even though this behavior could be avoided by increasing the prediction horizon, it would still appear for obstacles larger than the rollout distance.
Therefore, leaving the dead-end detection task to a high-level planning algorithm providing the reference states $\x^{\text{ref}}_{j}$ might prove more suitable.


\subsection{Discussion and future work} \label{section:discussion}
Future work extending the proposed method includes using the so-called \textit{Delay compensation by prediction}~\cite{nocse_diehl_2022}.
The time of the computation of each iteration is consistent thanks to a fixed number of computations performed per iteration.
Therefore, instead of addressing the problem starting at the current state, we can forward-simulate the system and start the optimization at a state at which the system will be when we would have computed the optimization iteration.

The selection of the noise parameters $\Sigma$ could also be extended.
For now, the values were hand-tuned for fast flying and would not be suitable for, e.g., hovering.
Introducing a method to dynamically set the disturbance noise based on the reference variance could be more suitable, adjusting the noise variance during the flight.

Most notably, extensive real-world testing of avoiding (even dynamic) obstacles is planned.
For that, known obstacles need to be placed in the testing arena and inserted into the collision checking module, or the platform needs to be extended to include a sensor able to detect obstacles.

\section{Conclusion} \label{section:conclusion}
In this work, we presented a model predictive planning and control methodology called Model Predictive Path Integral~(MPPI) designed to control a drone along a given trajectory, while allowing the use of arbitrary cost functions and constraints.
We proposed a method that respects low-level constraints imposed by the motor thrust limits of the UAV, while using a nonlinear model of the system dynamics to predict the future dynamic states.
Moreover, we discussed the shortcomings of the traditional Euclidean metric used for tracking a reference trajectory considering the rotation part of the state and proposed a more suitable approach.

Validation of the complete approach was conducted using a UAV both in simulation and in outdoor flight using a real drone.
We showed that our proposed and implemented MPPI controller is able to control a drone at the level of commanded angular velocities and collective thrust while running all the needed computations at $\SI{100}{\hertz}$ online and onboard.
To the best of our knowledge, this is the first deployment of MPPI on a real drone without relying on external hardware, such as an offboard PC with a large GPU.

We compared our controller with two other controllers used for reference tracking, namely MPC and SE(3).
The results show that the proposed method consistently outperforms the original MPC algorithm, achieving around $\SI{60}{\percent}$ better tracking error in all test scenarios.
Although unable to reach the performance of the SE(3) controller, it has the ability to include obstacles in the control task, which is impossible in the SE(3) controller and greatly limited in the existing MPC controllers.
The ability of the MPPI controller to avoid non-convex obstacles by using a general collision detection module is demonstrated in two experimental setups, tracking a reference at up to $\SI{44}{\kilo\meter\per\hour}$ and acceleration close to $\SI{20}{\meter\per\second\squared}$ while consistently avoiding obstacles.
This proves the biggest strength of the proposed method, which lies in the ability to handle non-convex and even non-differentiable cost functions, therefore enabling the general collision avoidance mentioned.
\vspace{-0.5em}

\input{reference.bbl}

\end{document}